# Normalization and the Representation of Nonmonotonic Knowledge in the Theory of Evidence

**Ronald R. Yager**
Machine Intelligence Institute, Iona College, New Rochelle, N.Y. 10801
and
NASA, Ames Research Center, Moffett Field, Ca 94035

**ABSTRACT**
*We discuss the Dempster-Shafer theory of evidence. We introduce a concept of monotonicity which is related to the diminution of the range between belief and plausibility. We show that the accumulation of knowledge in this framework exhibits a nonmonotonic property. We show how the belief structure can be used to represent typical or commonsense knowledge.*

## 1. Introduction

What appears to be a central characteristic of commonsense knowledge is the possibility of nonmonotonicity. By this we mean that the addition of knowledge may cause us to withdraw some previously made inference.

In this paper, we look at the theory of evidence (Demp 1967 & Shafer 1976) and particularly the process of normalization. We show that this process introduces an inherent nonmonotonicity into the Dempster-Shafer framework. This inherent nonmonotonicity, we feel, makes this representation scheme a suitable one for the representation of default knowledge. We then show how the belief structure provides a formalism for representing commonsense knowledge.

## 2. Information and Monotonicity in the Theory of Evidence

In this section, we introduce a number of issues related to the information contained in a belief structure and provide a concept of monotonicity for the belief structures.

Assume m is a belief structure on the set X. Basically the information contained in a belief structure consists of knowledge about the probabilities of events in X. This view is very much in the spirit of Dempster's original work. We note that the essential feature of the belief structure is that the information it contains about the probabilities is generally imprecise (lacks specificity). Given the belief structure m for any subset A of X, the probability of A is known to satisfy the following inequality.

$$Bel(A) \leq Prob(A) \leq Pl(A).$$

For a given event A, we can use the term

$$r_A = Pl(A) - Bel(A)$$

to measure the uncertainty associated with our knowledge of the probability of A. If $r_A = 0$ then we exactly know the probability of A. Shafer calls a belief structure in which $r_A = 0$, for all A, a Bayesian belief structure. This type of structure requires that the focal elements be all singletons. It is essentially the classical probability structure.

In order to motivate and justify the work that follows, we make the following simple observation. Assume

$$a_1 \leq Prob(A) \leq b_1$$

then it naturally follows that

$$a_2 \leq Prob(A) \leq b_2$$

for any $a_2 \leq a_1$ and $b_2 \geq b_1$.

Put another, way given that $Prob(A) \in [a_1, b_1]$ we can infer that $Prob(A) \in [a_2, b_2]$. We note that the second interval is wider, less specific, but still valid. In the above we note that $[a_1, b_1] \subset [a_2, b_2]$.

This observation provides the basis for the introduction of an inference among belief structures which was introduced by Yager (1986).

Assume $m_1$ and $m_2$ are two belief structures on X such that for every subset A of X the following condition is satisfied

$$[Bel_1(A), Pl_1(A)] \subset [Bel_2(A), Pl_2(A)]$$

If the above condition holds then it follows that knowledge that $m_1$ is true allows us to infer that $m_2$ is true. A more general formulation of this result is possible.

**Def.:** Assume $m_1$ is a belief structure in which $A_1, ... A_p$ are the focal elements with weights $m(A_i) = a_i$. Let $m_2$ be another belief structure in which the focal elements can be represented as

$$B_{11}, B_{12}, ... B_{1n_1}, B_{21}, B_{22}, ... B_{2n_2}, ... B_{p1}, B_{p2}, ... B_{pn_p}$$

where

$$m_2(B_{ij}) = b_{ij}.$$

In addition assume that for each $i = 1, .....$ p the following two conditions are satisfied

1. $A_i \subset B_{ij}$ for all $j = 1, ... n_i$

2. $\sum_{j=1}^{n} m(B_{ij}) = a_i$

In this situation we say that $m_1$ entails $m_2$ and denote this as

$$m_1 \subset m_2.$$

Yager (1986) proves the following result.
**Theorem:** If

$$m_1 \subset m_2.$$

then for every subset A

$$[Bel_1(A), Pl_1(A)] \subset [Bel_2(A), Pl_2(A)]$$

As a result of this theorem we can see that if $m_1 \subset m_2$ then the knowledge that $m_1$ is a valid description of the world allows us to infer that $m_2$ is also true. We can see that if $m_1 \subset m_2$ then $m_1 \vdash m_2$. We can say that if $m_1 \subset m_2$ $m_1$ is a more specific representation.

We should note that the process of gaining knowledge in the D-S framework can be seen as obtaining more specific belief functions as the representation of our knowledge.

In logic the concept of monotonicity plays a fundamental role. It essentially is a manifestation of the fact that the gaining of knowledge allows us to know more about the world.

Assume $P_1, P_2, ... P_n$ are a collection of propositions from which we can deduce the proposition H, we denote this as

$$(P_1, P_2, ... P_n) \vdash H$$

The situation is considered to be monotonic if the addition of any proposition, $p_{n+1}$, still allows us to infer H, that is

$$(P_1, P_2, \ P_n, P_{n+1}) \vdash H$$

The environment is called non-monotonic if the addition of some proposition may require us to withdraw the validity of H.

We are now in a position to introduce an analogous concept of monotonicity in the Dempster-Shafer framework. In the Dempster-Shafer framework the role of propositions are played by belief structures. Assume $m_1, m_2, \ldots m_q$ are a collection of belief structures. The process of reasoning in this framework consists of the aggregation (conjunction) of pieces of evidence. In particular if

$$m^* = m_1 \cap m_2 \cap \ldots \cap m_q$$

then

$$(m_1, m_2, \ldots m_q) \vdash m$$

where m is any belief structure such that

$$m^* \subset m.$$

Thus in the D-S framework we can infer any belief structure which contains (is entailed by) the intersection of the constituent belief structures. We shall say that the situation is monotonic if for any additional $m_{q+1}$, the conjunction

$$m_1 \cap m_2 \ldots m_q \cap m_{q+1} = m^+$$

satisfies

$$m^+ \subset m^* \ \ (m^* \cap m_{q+1} \subset m^*)$$

This concept of monotonicity is based upon the fact that, if $m^+ \subset m^*$ then from $m^+$ we can use the entailment principle to infer any belief structure inferred by $m^*$.

We should note that the view taken in this Dempster-Shafer approach is essentially a generalization of the model theoretic approach in logic. In that view we associate with any proposition $P_i$ a set $S_i$ of possible worlds that are true under $P_i$. Then

$$S^* = S_1 \cap \ldots S_q$$

are the worlds acceptable to all the propositions. A proposition H is inferable if the set of possible worlds, $S_H$, that make H true satisfies

$$S^* \subset S_H.$$

## 3. Normalization and Non-Monotonicity

In this section we shall investigate the reasoning process in the Dempster-Shafer theory, the conjunction of pieces of evidence, in regards to its satisfying the previously described monotonicity condition. We shall show that in general monotonicity is not always guaranteed in the Dempster-Shafer framework. More specifically, when the aggregation of evidence has conflicts, focal elements that have null intersection, then a nonmonotonicity is introduced. On the other hand if there is no conflicts amongst the evidence, then monotonicity is guaranteed. Thus it is generally not the case that

$$m_1 \cap m_2 \subset m_1,$$

only when there is no conflicts does the result hold.

The following example will help illustrate the issues. Assume that $m_1$ and $m_2$ are two belief structures such that

$$m_1(A) = \alpha \quad m_2(B) = \beta$$

$$m_1(X) = 1 - \alpha \quad m_2(X) = 1 - \beta$$

Using Dempster's rule if $A \cap B \neq \Phi$ the aggregation of these two pieces of evidence results in the belief structure m where

| Focal element | Weight |
|---|---|
| $A \cap B$ | $\alpha\beta$ |
| A | $\alpha\bar{\beta}$ |
| B | $\bar{\alpha}\beta$ |
| X | $\bar{\alpha}\bar{\beta}$ |

We note in this case that

$$m_1 \cap m_2 \subset m_1$$

as well as

$$m_1 \cap m_2 \subset m_2$$

The fact that

$$m_1 \cap m_2 \subset m_1$$

follows from the fact that $A \cap B$ and A are both contained in A and have a sum of weights equal $\alpha$. In addition both B and X are contained in X and sum to $\bar{\alpha}$.

More generally we can prove the following theorem Yager (1986):

**Theorem:** Assume $m_1$ and $m_2$ are two belief structures with focal elements $A_1, ... A_p$ and $B_1, ... B_q$. If for all $A_i$ and $B_j$, $A_i \cap B_j \neq \Phi$ then

$$m_1 \cap m_2 \subset m_1$$

and $m_1 \cap m_2 \subset m_2$.

Thus we see that if no conflicts arise the reasoning process is monotonic. An implication of this observation is that, as we obtain more information, in terms of additional belief structures, which doesn't conflict with what we already have then our overall knowledge increases. This follows since our ranges, the $r_A$'s get smaller. In the case in which conflicts arise between focal elements the situation becomes different. If we consider the previous example but with the case in which $A \cap B = \Phi$ then we get

| Focal element | Weight |
|---|---|
| A | $\alpha\bar{\beta}/1-\alpha\beta$ |
| B | $\bar{\alpha}\beta/1-\alpha\beta$ |
| X | $\bar{\alpha}\bar{\beta}/1-\alpha\beta$ |

In this case it is not necessarily the case that $m_1 \cap m_2 \subset m_1$. This case can be seen as follows. We first note that

$$Pl_1(A) = 1$$

$$Bel_1(A) = \alpha$$

thus

$$Prob_1(A) \in [\alpha, 1]$$

In $m = m_1 \cap m_2$, since $A \cap B = \Phi$ then

$$Pl(A) = (1-\bar{\alpha}\beta)/(1-\alpha\beta) = a$$

thus
$$\text{Bel}(A) = \alpha\bar{\beta}/1-\alpha\beta = b$$

$$\text{Prob}(A) \in [b,a]$$

However since

$$b = \alpha\bar{\beta}/1-\alpha\beta = \alpha(1-\beta)/1-\alpha\beta < \alpha$$

then

$$\text{Bel}(A) < \alpha$$

thus

$$[\text{Bel}(A), \text{Pl}(A)] \not\subset [\text{Bel}_1(A), \text{Pl}_1(A)]$$

The process exhibits a nonmonotonicity. Thus it appears the introduction of conflict and the requirement for normalization introduces a nonmonotonicity into the process.

As a first step in reducing this non-monotonicity we may try to use a different process for normalization then that used in Dempster's rule. We note that Dempster's rule essentially allocates the weights in the conflict set proportionately amongst the non conflicting focal elements.

Dubois and Prade(1988) provide a comprehensive discussion of the issue of normalization in the aggregation of belief structures. In that paper they discuss a number of alternative procedures for aggregation. We shall look at a number of these and see their effect on the monotonicity issue.

One alternative approach to normalization was suggested by Yager(1987). In this approach instead of proportionately distributing the conflict amongst all the focal elements we give it all to the base set X. Thus using this approach we get when $A \cap B = \Phi$

$$m_1 \cap m_2 = m^*$$

where in this case

$$m^*(A) = \alpha\bar{\beta}$$
$$m^*(B) = \bar{\alpha}\beta$$
$$m^*(X) = \overline{\alpha\beta} + \alpha\beta$$

Hence in this case

$$\text{Pl}^*(A) = \alpha\bar{\beta} + \overline{\alpha\beta} + \alpha\beta = 1 - \bar{\alpha}\beta = a^*$$
$$\text{Bel}^*(A) = \alpha\bar{\beta} = b^*$$

Again we note that

$$b^* = \alpha\bar{\beta} < \alpha$$

and hence

$$[\text{Bel}^*(A), \text{Pl}^*(A)] \not\subset [\text{Bel}_1(A), \text{Pl}_1(B)]$$

Thus the nonmonotonicity has not been eliminated. A second approach, one suggested by Dubois and Prade, is to replace conflict sets with the union of the two sets. Thus if $A \cap B = \Phi$ then use $A \cup B$. In this case we get

$$m_1 \cap m_2 = m^+$$

where

$$m^+(A \cup B) = \alpha\beta$$
$$m^+(A) = \alpha\bar{\beta}$$
$$m^+(B) = \bar{\alpha}\beta$$
$$m^+(X) = \overline{\alpha\beta}$$

In this case we see that

$$Pl^+(A) = 1 - \overline{\alpha}\beta = a^+$$
$$Bel^+(A) = \alpha\overline{\beta} = b^+$$

Since $b^+ = b^*$ and $a^+ = a^*$ the situation is the same as in the previous method and the nonmonotonicity still exists.

A third approach consists of applying a type of weighted averaging which essentially corresponds to a discounting. In this approach we calculate

$$m° = cm_1 + (1 - c)m_2$$

It can be shown that this approach still doesn't reduce the nonmonotonicity.

Is there any procedure we can use that will reduce the nonmonotonicity? Essentially we see that the issue is that we must assign the conflict weight $\alpha\beta$ in a manner which brings up the $Bel(A)$ to $\alpha$. Thus we must assign $\alpha\beta$ to some set D such that $D \subset A$, because in that case

$$Bel(A) = \alpha\overline{\beta} + \alpha\beta = \alpha$$

But must also do it in a way that $D \subset B$ so that we keep the monotonicity with respect to $m_2$, this requires that $D \subset B$. Since $B \cap A = \Phi$ the only possible value for D is the null set. Thus one possibility is to forgo the process of normalization. In this case we get

$$m^\nabla = m_1 \cap m_2$$

where

$$m^\nabla(\Phi) = \alpha\beta$$
$$m^\nabla(A) = \alpha\overline{\beta}$$
$$m^\nabla(B) = \overline{\alpha}\beta$$
$$m^\nabla(X) \; \overline{\alpha}\overline{\beta}$$

In this case we get that

$$m^\nabla \subset m_1$$

and thus the situation is monotonic. However the price we pay for this is that

$$Pl(X) \neq 1.$$

We can consider a more general rule for the intersection of focal elements in the Dempster's rule. Assume A and B are two focal elements from $m_1$ and $m_2$ then we define their conjunction as

$$D = (A \cap B) \cup (1 - Poss(B/A)) * F(A,B))$$

In the above the function $F(A,B)$ results in some subset of X. In all case when $Poss(B/A) = 1$ we get $A \cap B$. When $Poss(B/A) = 0$ the value depends upon the choice of F.

In particular if we set

$$F(A,B) = A \cup B$$

then when $Poss(B/A) = 0$ we get that

$$D = A \cup B$$

which is the suggestion of Dubois & Prade (1988).

If we let

$$F(A,B) = \overline{A} \cup \overline{B}$$

then when $A \cap B = \Phi$ we get

$$F(A,B) = X$$

which is the suggestion of Yager (1987).

If we let

$$F(A,B) = A$$

399

then when $A \cap B = \Phi$ we get
$$D = A.$$
The use of this last rule guarantees, even if conflicts arise, that
$$m_1 \cap m_2 \subset m_1$$
but not
$$m_1 \cap m_2 \subset m_2.$$
This last approach introduces an idea of priority. It can be seen as a form of discounting, where $m_2$ is discounted if it conflicts with $m_1$ otherwise we don't discount it.

## 4. Default Knowledge in Belief Structures

In this section we show that the appearance of the nonmonotonicity rather than being a problem provides a natural facility for introducing default knowledge in the framework of belief structures.

Assume V is a variable that takes its value in the set X. Assume we have two pieces of knowledge with respect to V. The first an absolute and the second a typical piece of knowledge:
$$P_1: \text{V is A}$$
$$P_2: \text{typically V is B}.$$

As discussed by Yager(1987b), essentially what we desire to happen in this case is that, if $A \cap B \neq \Phi$ we obtain $A \cap B$ as our inference. If they don't intersect we discount the default knowledge and are left with A. We shall see that the reasoning mechanism in the theory of evidence essentially provides this capability.

Yager(To Appear) has suggested that one can represent a statement like
$$\text{typically V is B}$$
by a belief structure
$$\text{V is } m_2$$
where
$$m_2(B) = \alpha$$
$$m_2(X) = 1 - \alpha$$
where $\alpha$ is close to one. The idea being that by typical knowledge we are saying that there is a high probability that V lies in the set B but some small chance it can be anywhere.

Thus the above knowledge base can be represented in the Dempster-Shafer framework as two belief structures $m_1$ and $m_2$. Where
$$m_1(A) = 1 \qquad m_2(B) = \alpha$$
$$m_2(X) = 1 - \alpha$$
Taking the conjunction of these two pieces of information we get
$$m = m_1 \cap m_2.$$
If $A \cap B \neq \Phi$ then we get
$$m(A \cap B) = \alpha$$
$$m(A) = 1 - \alpha$$
In this case
$$Pl(A) = Bel(A) = Prob(A) = 1$$

400

thus we are certain that V lies in A. In addition
$$Pl(A \cap B) = 1$$
$$Bel(A \cap B) = \alpha$$
thus
$$\alpha \leq Prob(A \cap B) \leq 1$$
So that we get, as desired, that there is a high probability that V lies in $A \cap B$.

On the other hand if $A \cap B = \Phi$ and we use Dempster's normalization process we get
$$m(A) = 1.$$
Thus we see that we get the desired result that the typical knowledge is completely discounted when a conflict with absolute knowledge arises.

## 5. Priorities and Strengths of Defaults

We next consider the case in which we have two pieces of default knowledge:
$P_1$: typically V is A
$P_2$: typically V is B

In addressing this problem by one of the logical extension type methods we are faced with the issue of priority amongst the default rules. In particular if one of the default rules has a higher priority it is introduced first in the deduction process. If $A \cap B \neq \Phi$ no problems arise and we infer $A \cap B$. If $A \cap B = \Phi$ and if $P_1$ is considered to have a higher priority then we should infer A. While if $P_2$ is considered to have a higher priority we should infer B. If $P_1$ and $P_2$ are considered to have some priority we should get a different result which is essentially the union of the two. Subsequently we shall see that the role of priority in the D-S framework is played by the probability assigned to the default set in each case.

We can represent the above knowledge base by two belief structures $m_1$ and $m_2$ where
$$m_1(A) = \alpha \quad m_2(B) = \beta$$
$$m_1(X) = 1 - \alpha \quad m_2(X) = 1 - \beta$$

In this representation we can see that $\alpha$ essentially describes the strength of the first default rule since $Bel(A) = \alpha$. That is $\alpha$ measures how certain we are that A holds.

In the above example our inferred belief structure is
$$m = m_1 \cap m_2$$

If $A \cap B \neq \Phi$ then we get
$$m(A \cap B) = \alpha\beta$$
$$m(A) = \alpha\bar{\beta}$$
$$m(B) = \bar{\alpha}\beta$$
$$m(X) = \overline{\alpha\beta}$$

In this case
$$Pl(A \cap B) = 1$$
$$Bel(A \cap B) = \alpha\beta$$
thus
$$Prob(A \cap B) \geq \alpha\beta$$

Since both $\alpha$ and $\beta$ are very close to one this result essentially says that the inferred value for V is $A \cap B$. Since both A and B have small probabilities we can use the entailment principle to replace the above by

$$m(A \cap B) = \alpha\beta$$
$$m(X) = 1 - \alpha\beta$$

This essentially says that we get typically $A \cap B$ with a little less strength.

If $A \cap B = \Phi$ then we get

$$m(A) = \alpha\bar{\beta}/(1 - \alpha\beta)$$
$$m(B) = \bar{\alpha}\beta/(1 - \alpha\beta)$$
$$m(X) = \bar{\alpha}\bar{\beta}/(1 - \alpha\beta)$$

Let us consider the case in which $P_1$ is a stronger default rule then $P_2$, $\alpha > \beta$. This essentially corresponds to $P_1$ having a priority over $P_2$. In the above since

$$Bel(A) = \alpha\bar{\beta}/(1 - \alpha\beta)$$
$$Bel(B) = \bar{\alpha}\beta/(1 - \alpha\beta)$$

and with $\alpha > \beta$ it follows that $\alpha\bar{\beta} > \bar{\alpha}\beta$ we see that $Bel(A) > Bel(B)$. The actual distinction depends upon the values for $\alpha$ and $\beta$, the strengths of the defaults.

Example: If $\alpha = .99$ and $\beta = .9$ then

$$m(A) = .9$$
$$m(B) = .08$$
$$m(X) = .02$$

In this case

$$.9 \leq Prob(A) \leq .92$$

and

$$.08 \leq Prob(B) \leq .1$$

Thus there is a high probability that V lies in A and a small chance that V lies in B. In addition

$$Prob(A \cup B) \geq .98$$

In the case when $P_1$ and $P_2$ have the same strength (priority) then $\alpha = \beta$. In this case

$$m(A) = \alpha\bar{\alpha}/(1 - \alpha^2)$$
$$m(B) = \alpha\bar{\alpha}/(1 - \alpha^2)$$
$$m(X) = \bar{\alpha}\bar{\alpha}/(1 - \alpha^2)$$

In this case the belief and plausibility of A and B are equal.

Example: If $\alpha = .95$ then

$$m(A) = .49$$
$$m(B) = .49$$
$$m(X) = .02$$

In this case Prob(A) and Prob(B) are contained in (49,51). Which essentially says that we are about 50 percent certain of each. From a pragmatic point of view if two rules are conflicting and are equal priority it would appear that we should not assign to high a strength to each. Thus if we let $\alpha = .5$ we get

$$m(A) = 1/3$$

In this case
$$m(B) = 1/3$$
$$m(X) = 1/3$$
$$1/3 \leq \text{Prob}(A) \leq 2/3$$
and
$$2/3 \leq \text{Prob}(A \cup B) \leq 1.$$

Assume we have a proposition

typically V is A

to which we assign a strength $\alpha$. Therefore
$$m(A) = \alpha$$
$$m(X) = 1 - \alpha$$
From the entailment principle we can infer
$$m^*(A) = \alpha_1$$
$$m^*(X) = 1 - \alpha_1$$
where
$$\alpha_1 < \alpha.$$

Essentially what this implies is that we can always hedge, make less the degree of typicality, and still be correct.

## 6. References

Dempster, A.P., (1967) ,"Upper and lower probabilities induced by a multivalued mapping," Ann. Math. Statistics 38, 325-339.

Dubois, D. & Prade, H., (1988), "Representation and combination of uncertainty with belief functions and possibility measures," Computational Intelligence 4, 244-264.

Shafer, G., (1976), A Mathematical Theory of Evidence, Princeton University Press: Princeton, N.J..

Yager, R.R., (1986), "The entailment principle for Dempster-Shafer granules," Int. J. of Intelligent Systems 1, 247-262.

Yager, R.R., (1987), "On the Dempster-Shafer framework and new combination rules," Information Sciences 41, 93-137.

Yager, R.R., (1987b) "Using approximate reasoning to represent default knowledge," Artificial Intelligence 31, 99-112.

Yager, R.R., (To Appear), "On usual values in commonsense reasoning," Fuzzy Sets and Systems